# Soft Robotics for Search and Rescue: Advancements, Challenges, and Future Directions

Abhishek Sebastian[1][0000-0002-3421-1450]

[1] Division of Robotics, Abhira

**Abstract.** Soft robotics has emerged as a transformative technology in Search and Rescue (SAR) operations, addressing challenges in navigating complex, hazardous environments that often limit traditional rigid robots. This paper critically examines advancements in soft robotic technologies tailored for SAR applications, focusing on their unique capabilities in adaptability, safety, and efficiency. By leveraging bio-inspired designs, flexible materials, and advanced locomotion mechanisms, such as crawling, rolling, and shape morphing, soft robots demonstrate exceptional potential in disaster scenarios. However, significant barriers persist, including material durability, power inefficiency, sensor integration, and control complexity. This comprehensive review highlights the current state of soft robotics in SAR, discusses simulation methodologies and hardware validations, and introduces performance metrics essential for their evaluation. By bridging the gap between theoretical advancements and practical deployment, this study underscores the potential of soft robotic systems to revolutionize SAR missions and advocates for continued interdisciplinary innovation to overcome existing limitations.

**Keywords:** Soft Robotics, Search and Rescue (SAR), Disaster Response.

## 1    INTRODUCTION

Search and Rescue (SAR) operations aim to locate, assist, and rescue victims trapped or endangered in disaster scenarios such as earthquakes, floods, building collapses, and other catastrophic events [1]. These operations are characterized by unpredictable, hazardous, and often dynamic environments that require rapid response and innovative tools to ensure the safety and effectiveness of the rescue process. Over the years, robotics has emerged as a critical component of SAR, augmenting human capabilities in scenarios that are dangerous, inaccessible, or otherwise beyond human capability. Traditional rigid robots, while effective in certain applications, frequently encounter significant limitations when faced with the complexities of SAR environments. Their rigid structures often struggle with navigating narrow passages, unstable rubble, or uneven terrains, which are common in disaster zones [2].

In contrast, soft robotics has gained attention as a transformative approach, employing flexible, deformable, and adaptive materials inspired by biological systems [3][4]. Unlike their rigid counterparts, soft robots excel in their ability to safely interact with fragile structures without causing additional damage [5]. Their exceptional adaptability enables them to access confined or irregular spaces [6] and traverse uneven or unpredictable surfaces [7]. For instance, bio-inspired designs such as tentacle-like or worm-like movements allow soft robots to mimic natural forms of locomotion, giving them the ability to maneuver effectively through complex disaster environments [8]. This flexibility not only enhances their physical capabilities but also ensures greater safety in human-robot interactions, a critical requirement during rescue operations.

Despite their potential, the adoption of soft robotics in SAR is not without challenges. Key issues include the limited durability of soft materials when exposed to harsh and abrasive conditions, the power inefficiency of soft actuators during extended operations, and the inherent complexity in embedding sensors and actuators into flexible and deformable bodies [10][11]. Furthermore, designing control algorithms that can reliably guide soft robots through chaotic, unstructured, and constantly changing disaster environments remains a significant technical hurdle [12]. Addressing these challenges requires a multidisciplinary research effort, incorporating advances in material science, robotics, artificial intelligence, and sensor technologies to fully realize the potential of soft robotics in SAR applications.

There have been a variety of reviews and surveys conducted in the fields of bio-inspired robotics [13], soft material modeling [14], advanced control systems [15], and machine learning applications in robotics [16], highlighting significant progress and advancements in these domains. These reviews have shed light on the potential of soft robotics in diverse applications, ranging from industrial automation to biomedical devices. However, a distinct gap exists in the literature specifically addressing the application of soft robots in Search and Rescue (SAR) operations. SAR scenarios present unique and demanding challenges, such as navigating unpredictable terrains filled with debris, safely interacting with survivors in fragile environments, and performing critical, high-stakes tasks under time constraints where reliability is paramount.

This paper reviews soft robotic technologies for Search and Rescue (SAR) applications, examining their current capabilities, limitations, and future directions. By bridging the gap between theoretical advancements and real-world deployment, it highlights the transformative potential of soft robotics in disaster response and serves as a resource for researchers and practitioners in this field. The discussion is structured around key aspects, including disaster characteristics, the need for soft robotics in SAR, and the unique advantages these systems offer in navigating complex and hazardous environments.

The review explores core soft robotic technologies, beginning with material selection, analyzing elastomers, hydrogels, and shape-memory alloys for their SAR applications. It also delves into advanced locomotion mechanisms such as crawling, rolling, and shape morphing, evaluating their effectiveness in disaster scenarios. Additionally, the paper examines sensing capabilities in harsh conditions, addressing challenges in sensor integration and showcasing cutting-edge developments that enhance perception in flexible robotic systems.

To assess performance, the paper investigates simulation techniques, hardware tests, and proposes a framework of standards and performance metrics for SAR robots. It also discusses fundamental challenges, including terrain adaptability and the complexities of modeling and controlling soft robots. Concluding with future directions, the paper underscores the need for continued innovation and interdisciplinary collaboration to advance soft robotics in SAR applications.

## 2 DISASTER CHARACTERISTICS AND IMPACT ON ROBOTICS

Disasters, whether natural or manmade, are complex events that can cause widespread damage, disrupt communities, and challenge rescue efforts (Figure 1.). Understanding their categories and phases is critical for designing and deploying effective robotic solutions. This section explores the two main types of disasters — natural and manmade — and the unique challenges they present to robotics in Search and Rescue (SAR) operations.

Natural disasters originate from environmental or geological processes. These include earthquakes, floods, hurricanes, tsunamis, volcanic eruptions, and landslides [17]. Each of these disasters has distinct characteristics that shape the environment in which rescue robots must operate [18]. For instance, earthquakes often result in collapsed buildings, creating rubble-filled terrains with narrow spaces and unstable surfaces [19]. Floods, on the other hand, submerge vast areas under water, rendering conventional rescue methods ineffective and requiring amphibious robotic systems to navigate [20]. Hurricanes combine strong winds, heavy rain, and debris, often necessitating aerial robots for reconnaissance in areas inaccessible by land [21].

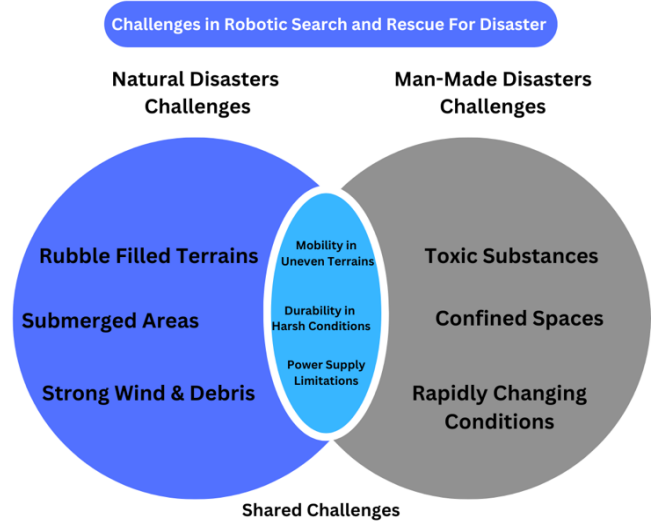

**Fig. 1.** Challenges in Robotics Search and Rescue for Disasters.

In natural disaster scenarios, robots face several significant challenges. The terrain is often unpredictable, with obstacles such as debris, mud, or water complicating movement. The environment may be hazardous, exposing robots to extreme temperatures, corrosive materials, or high pressure, depending on the type of disaster [22]. Additionally, limited visibility caused by smoke, water, or dust can hinder sensing and navigation, making robust sensor systems essential for successful operation [23].

Manmade disasters result from human activities, either accidental or deliberate. Examples include industrial fires, structural collapses, hazardous material spills, and terrorist attacks [24]. While these events may not always cover as large an area as natural disasters, they often pose a concentrated level of danger that is just as difficult to navigate. For example, structural collapses in urban environments can create unstable, debris-filled zones like those seen in earthquakes [19]. Fires introduce extreme heat and smoke, demanding materials and sensors that can withstand such conditions. Hazardous spills, such as chemical or radioactive leaks, add the challenge of avoiding contamination, both for the robot and for the rescue team it supports [26].

Robots operating in manmade disasters face additional challenges compared to natural disasters. In many cases, the presence of toxic or flammable substances means the robots must be non-conductive or explosion-proof. Navigation in confined spaces, such as collapsed tunnels or building interiors, requires robots to be highly maneuverable and equipped with advanced tactile and environmental sensors [27]. Moreover, the dynamic nature of these disasters, where conditions can change rapidly, necessitates quick adaptability and reliable communication systems to relay real-time data to human operators [28].

Both natural and manmade disasters share a key set of challenges that influence the design of SAR robots. Mobility is a primary concern, as disaster zones are rarely uniform or predictable. Robots must be capable of overcoming uneven surfaces, climbing debris, and squeezing through narrow gaps [29][30]. Durability is another critical factor, as robots need to withstand the harsh physical and environmental conditions present in these settings [31]. Power supply is equally vital; the inability to recharge or refuel in remote or dangerous locations can limit a robot's effectiveness [32].

## 3   TYPES OF SOFT ROBOTS USED IN SAR

Soft robots are categorized into six primary types based on their actuation mechanisms and materials (Table 1.). Each type offers unique capabilities tailored to specific applications in challenging environments, such as disaster zones and underwater rescues. The table below highlights their key features, potential applications, and associated challenges in real-world scenarios.

**Table 1.** Types of Different Soft Robots

| Type | Description | Applications | Challenges |
| --- | --- | --- | --- |
| Pneumatic Soft Robots | Operate by inflating or deflating internal chambers. | Navigating rubble, inspecting tight gaps in collapsed structures. | Limited portability due to air supply requirements, energy-intensive operation. |
| Hydrogel-Based Soft Robots | Constructed from water-absorbent materials. | Underwater rescues, chemical sensing in hazardous conditions. | Susceptibility to degradation in polluted or high-temperature environments. |
| Origami-Inspired Soft Robots | Enable robots to fold and unfold their structure. | Expanding pathways, stabilizing structures in rubble. | High precision required for folding mechanisms, reduced durability under load. |
| Magnetically Actuated Soft Robots | Utilize embedded magnetic particles that respond to external magnetic fields. | Navigating collapsed structures, inspecting pipes or voids. | Dependency on external magnetic field generators, reduced efficiency in cluttered environments. |
| Soft Robots with Shape-Memory Materials | Can alter their form when exposed to certain stimuli. | Penetrating compact debris, supporting fragile areas. | Energy-intensive transformation processes, material fatigue over repeated use. |

## 4   ADVANTAGES AND LIMITATIONS OF SOFT ROBOTS IN CHALLENGING ENVIRONMENTS COMPARED TO RIGID ROBOTS.

One of the primary benefits of soft robots in SAR missions is their exceptional adaptability to challenging environments. Constructed from flexible materials like elasto-mers and hydrogels, soft robots can deform, bend, and stretch, enabling them to navigate through tight spaces and conform to irregular surfaces [33][34][35][36]. This flexibility allows them to traverse debris, rubble, and other obstacles commonly found in disaster zones, reaching areas that might be inaccessible to rigid robots or human rescuers [39].

Soft robots also offer safety benefits, particularly when operating near humans. Their compliant nature reduces the risk of injury during physical interactions, making them suitable for tasks that involve direct contact with survivors [37][38]. This characteristic is crucial in SAR operations, where the safety of both victims and rescue personnel is paramount.

While soft robots bring unique advantages to Search and Rescue (SAR) operations, they are not without their drawbacks. In certain aspects, traditional rigid robots still outperform their soft counterparts, particularly in terms of strength, efficiency, and reliability. A detailed comparison (See Table 2.) between the two highlights key challenges that must be addressed for soft robots to reach their full potential in SAR applications.

Soft robots excel in environments where adaptability and safety are non-negotiable. Unlike rigid robots, their flexibility allows them to move through constrained spaces or over uneven debris where traditional designs would falter. Imagine a scenario where a survivor is trapped beneath a collapsed building. A soft robot, with its compliant structure, could navigate the tight gaps, gently interact with fragile surroundings, and deliver life-saving resources without exacerbating the situation. This capacity for safe, non-destructive operations is what sets soft robots apart, especially in high stakes rescue missions.

However, while their adaptability is their greatest strength, it also comes with significant trade-offs. Soft robots often struggle with heavy lifting or prolonged operations in abrasive environments. For example, the very materials that make them flexible — like elastomers and hydrogels — are prone to wear and tear when exposed to sharp debris or high temperatures. This is a critical limitation in scenarios where the robots might need to operate for extended periods or handle heavy loads, such as clearing rubble to access survivors.

**Table 2.** Comparative Analysis of Soft Robots and Rigid Robots w.r.t SAR

| Aspect | Soft Robots | Rigid Robots |
|---|---|---|
| Structural Strength | Limited structural strength; struggles with heavy lifting. | High strength, capable of handling heavy debris or obstacles. |
| Speed and Efficiency | Slower due to flexible movement and lower force generation. | Faster with more efficient locomotion mechanisms. |
| Durability | Prone to wear and tear in abrasive environments. | More resistant to sharp or rough surfaces. |
| Power Requirements | Limited power efficiency due to complex actuation mechanisms. | Better optimized for energy-efficient operations. |
| Sensor Integration | Challenging to embed rigid sensors into flexible materials. | Easier integration of robust and precise sensors. |

| Control Complexity | Requires advanced algorithms for real-time deformation control. | More straightforward control systems for pre-defined tasks. |
| --- | --- | --- |
| Autonomy | Limited gait diversity and autonomous operation capabilities. | More reliable autonomy and task-specific functionality. |
| Load Carrying Capacity | Poor weight-carrying ability; unsuitable for heavy loads. | Can handle higher payloads effectively. |
| Cost and Maintenance | High cost for custom materials and frequent maintenance. | Lower cost due to standardized materials and durability. |

Another major challenge is power efficiency. Soft robots typically rely on complex actuation mechanisms like pneumatics or hydraulics, which consume significant energy and limit operational duration. Unlike rigid robots, which can leverage optimized locomotion systems for energy-efficient movement, soft robots require novel solutions to enhance endurance. For SAR missions, where every second counts, the inability to operate continuously can be a serious drawback.

Sensor integration is another area that deserves attention. Embedding sensors into flexible materials is inherently challenging because traditional sensors are rigid and can compromise the robot's adaptability. While advancements in stretchable and flexible electronics are promising, these technologies are still in the early stages of development and often come with high costs. This hinders the ability of soft robots to perform critical tasks like detecting survivors' heat signatures or monitoring hazardous gases effectively.

Moreover, the control systems for soft robots add a layer of complexity. Unlike rigid robots, whose movements are predictable and can be pre-programmed, soft robots require advanced algorithms capable of managing real-time deformation. This means that the computational overhead is much higher, which can slow down decision-making in dynamic environments. For SAR missions, where conditions change rapidly, this lack of speed and reliability in control can limit the robot's effectiveness.

Finally, the issue of autonomy must be addressed. While soft robots can navigate unpredictable terrains, their ability to operate independently remains limited. Traditional rigid robots are better equipped with autonomous systems, allowing them to perform specific tasks like mapping disaster zones or identifying survivors without constant human intervention. Soft robots, on the other hand, often require manual guidance, reducing their practicality in large-scale or prolonged missions.

## 5 SOFT ROBOTIC TECHNOLOGIES FOR DISASTERS

Soft robotic technologies leverage materials and mechanisms inspired by biological systems to offer unparalleled adaptability, safety, and efficiency in disaster response. By combining innovative material selection, advanced locomotion mechanisms, and integrated sensing capabilities, these robots address the unique demands of Search and Rescue (SAR) operations. Their flexible nature allows them to traverse complex terrains, interact safely with survivors, and operate in confined spaces where traditional robots fall short.

This section explores soft robotic technologies' key components: flexible and durable materials, diverse locomotion techniques for adapting to varied disaster conditions, and effective sensing technologies for harsh environments. It highlights soft robotics' transformative potential in revolutionizing disaster response and the challenges needed to fully realize its capabilities.

## 5.1 Material selection

Material selection (Table 3.) is a critical aspect in the design and implementation of soft robotic technologies for disaster response [40]. The materials chosen must balance flexibility, durability, adaptability, and safety to ensure that the robots can navigate complex and hazardous environments effectively.

When selecting materials for soft robotic technologies in disaster response, several key considerations must be addressed to ensure optimal performance and reliability [41].

**Flexibility and compliance** are essential, allowing robots to conform to uneven surfaces and navigate through rubble, tight spaces, and irregular terrains with high elasticity [19]. **Durability and resilience** are crucial for withstanding harsh environmental conditions such as extreme temperatures, moisture, dust, and mechanical stresses without significant degradation. Utilizing **lightweight** materials enhances the robots' mobility and ease of deployment, which is vital for time-sensitive rescue operations [19][43]. **Actuation compatibility** ensures that the chosen materials work seamlessly with various actuators, including pneumatic, hydraulic, or shape-memory alloys, facilitating smooth and effective movements [43][44]. **Safety** is paramount, as materials must be non-toxic, non-abrasive, and safe for close interaction with human survivors. Additionally, **sensor integration** requires materials that can accommodate or interface with electronic components for navigation and hazard detection without compromising the robot's flexibility. Balancing these factors is essential for developing robust and efficient soft robots capable of operating effectively in disaster-stricken environments [43].

Material selection plays a pivotal role in the advancement of soft robotics, particularly in designing systems capable of adapting to complex and dynamic environments. Notable contributions in this domain emphasize the importance of flexibility, durability, and responsiveness in materials to achieve desired functionalities. Muralidharan et al. (2023) [100] introduced a bio-inspired soft jellyfish robot featuring a polyimide-based structure actuated by shape memory alloys (SMA). This innovative material choice provided high flexibility, thermal responsiveness, and durability, enabling the robot to mimic jellyfish locomotion for underwater applications, such as object detection using integrated camera and sonar systems. Similarly, Angelini et al. (2020) [54] developed the SoftHandler system, which incorporated flexible elastomers in its SoftDelta manipulator and Pisa/IIT SoftGripper. These materials offered the adaptability and compliance required to handle fragile and irregular objects, overcoming limitations of traditional industrial robots. Jin et al. (2016) [65] further advanced the field with their multi-layered smart modular structures (SMS), using SMA wires to actuate tentacle-like components. The SMS demonstrated exceptional flexibility, modularity, and load-bearing capabilities, enabling robots to adapt their morphology for terrestrial and underwater locomotion or delicate object handling.

Additional research underscores the potential of elastomeric materials in soft robotics. At Harvard Biodesign Lab [84], elastomeric matrices embedded with flexible reinforcements like cloth or paper have been used to fabricate soft fluidic actuators capable of diverse motions, including bending, twisting, and extension. These actuators leverage the inherent elasticity and moldability of silicone rubber, showcasing the versatility of elastomeric materials for creating compliant robotic systems.

Another critical innovation involves SMA-based actuators, where free-sliding SMA wires function as tendons to amplify bending angles and forces. This approach highlights the transformative potential of thermally responsive materials in enhancing actuator performance, particularly for tasks requiring large deformations and rapid actuation.

Collectively, these studies [47-100] demonstrate that the synergy between innovative materials and actuation technologies is foundational to the evolution of soft robotics. By tailoring material properties such as elasticity, thermal responsiveness, and mechanical durability, researchers are creating systems that excel in adaptability, safety, and efficiency. Whether through polyimide-based SMA actuators mimicking jellyfish, elastomeric matrices enabling complex motions, or thermoplastic components providing robust structural support, material selection remains at the core of developing next-generation soft robotic technologies capable of navigating and operating in diverse SAR environments.

**Table 3.** Different Materials used in Manufacturing Soft Robots for SAR

| Material | Advantages | Examples | Applications |
|---|---|---|---|
| Silicone Elastomers | **Highly Flexible:** Allows for significant deformation without damage | **PDMS (Polydimethylsiloxane):** Widely used due to its versatility.[47][48] | **Soft Grippers:** Handling fragile objects like debris.[53][54] |
|  | **Biocompatible:** Safe for human interaction. | **Ecoflex™ 00-30:** Known for its ultra-soft properties.[49][50][51][52][52] | **Flexible Actuators:** Enabling movement in confined spaces.[55] |
|  | **Transparent:** Facilitates integrated optical sensors. |  | **Compliant Structures:** Navigating |
| | | | through rubble and uneven terrains.[56] |
| Thermoplastic Elastomers [57] | **Durable:** Resistant to wear and tear. [58][59] | **Thermoplastic Polyurethane (TPU):** Offers excellent abrasion resistance.[61][62] | **Structural Components:** Providing both flexibility and strength in robot frames [57] |
| | **Ease of Manufacturing:** Can be molded or extruded.[57] | **SEBS (Styrene-Ethylene-Butylene-Styrene):** Known for its flexibility and toughness.[63] | **Protective Casings:** Shielding internal electronics from environmental hazards.[64] |
| | **Elasticity:** Combines rubber-like flexibility with plastic durability.[60] | | |
| Shape-Memory Alloys (SMA) [65][66] | **Adaptive:** Can revert to a pre-defined shape when heated. | **Nitinol:** Combines flexibility with high recovery force.[67][68][69] | **Actuators:** Powering movements in confined or hazardous environments.[65] |
| | **Lightweight:** Ideal for mobile robotic applications. | | **Adaptive Joints:** Providing precision in motion for navigation in tight spaces.[66] |
| Hydrogels [70][71][72] | **Self-Healing:** Capable of repairing minor damage autonomously. [72] | **Polyacrylamide Hydrogels:** Known for their stretchability and transparency.[73][74] | **Soft Sensors:** Detecting environmental changes or human touch during rescue.[77] |
| | **High Water Content:** Mimics soft tissue properties for gentle interaction. [72] | **Alginate-Based Hydrogels:** Biocompatible and easily modifiable.[75][76] | **Adhesive Pads:** Ensuring secure gripping on uneven surfaces.[78] |
| Liquid Crystal Polymers [79][80] | **Responsive:** Change properties based on environmental stimuli (e.g., temperature, light).[81] | **Vectran:** Known for its high strength and toughness.[82][83] | **Actuation Systems:** Enabling precise and adaptive motion in variable conditions.[83] |
| Polyurethane Foams [85][86] | **Lightweight:** Excellent for mobility.[87] | **Open-Cell Foams:** Provide compressibility and flexibility.[89] | **Shock Absorbers:** Protecting internal components during impact.[85] |
| | **Energy Absorbing:** Reduces damage during impacts.[88] | | **Insulation Layers:** Shielding sensitive electronics from extreme environmental conditions.[86] |
| Conductive Polymers [90] | **Electrically Conductive:** Supports integration with sensors and circuits.[91] | **PEDOT: PSS (Poly(3,4-ethylenedioxythiophene) polystyrene sulfonate):** Commonly used.[90] | **Embedded Sensors:** Detecting strain, pressure, or environmental hazards.[92] |
| | **Flexible and Lightweight:** Maintains mobility and ease of deployment.[90] | | **Flexible Circuits:** Enabling smooth electrical connectivity in moving components.[93] |
| Kevlar Composites [94] | **High Strength-to-Weight Ratio:** Excellent for protective layers.[95] | **Kevlar-Coated Fabrics:** Combines flexibility with toughness. [97] | **Reinforced Structures:** Providing mechanical integrity for |

| | | | disaster-response robots. [97] |
|---|---|---|---|
| | **Thermal and Chemical Resistance:** Operates under extreme conditions.[94] | | **Outer Shells:** Protecting from sharp debris and extreme environmental exposure.[97] |
| Polyimides [98] | **High Thermal Stability:** Withstands extreme temperatures. [98][99] | **Kapton:** Known for its heat resistance and electrical insulation properties. [100] | **Insulation Layers:** Protecting electronics from heat and electrical surges.[100] |
| | **Flexible and Durable:** Ideal for moving components.[99][96] | | **Flexible Substrates:** Supporting embedded circuits in actuating parts.[100] |

## 5.2 Locomotion Mechanisms.

Soft robotics employs diverse locomotion mechanisms —crawling, rolling, and shape morphing — each offering unique advantages (Figure 2.). Crawling excels in navigating confined spaces with flexible, deformable bodies. Rolling prioritizes speed and energy efficiency across diverse terrains. Shape morphing combines adaptability and multifunctionality, enabling dynamic responses to environmental challenges.

**a) Crawling locomotion**

Crawling locomotion in robotics pertains to a mode of movement that involves sequential contractions and expansions of the robotic body, akin to the motion observed in soft-bodied organisms such as worms and caterpillars. This form of locomotion is particularly advantageous for navigating constrained and complex environments, as it enables robots to adapt their shape and traverse obstacles with considerable flexibility. Nature serves as the primary inspiration for such mechanisms, where organisms utilize peristaltic waves and frictional variation to move efficiently across diverse terrains.

In 2013, Umedachi et al. pioneered the development of a highly deformable, 3D-printed soft robot inspired by soft-bodied animals [105]. This innovation facilitated crawling locomotion through variable friction legs integrated with a flexible body capable of substantial deformation, enabling it to traverse confined spaces. However, limitations such as slow locomotion speed and restricted adaptability to diverse terrains were noted in this design.

Subsequently, Pagano et al. in 2017 introduced a novel crawling robot that employed multi-stable origami structures [101]. By leveraging the unique characteristics of origami patterns, the robot achieved rapid shape transformations, thereby improving its locomotion efficiency. The multi-stable configurations facilitated quick transitions between shapes, enhancing its movement speed. Despite these advancements, challenges in precise control and maintaining stability across varied surfaces persisted.

Further progress was made in 2019 when Qin et al. designed a versatile soft crawling robot that demonstrated both rapid locomotion and adaptability to heterogeneous terrains [102]. This robot employed soft actuators mimicking muscular movements, significantly improving its crawling speed and ability to navigate uneven surfaces. Nevertheless, challenges related to energy efficiency and precise actuation control remained areas requiring further refinement.

In 2020, Sheng et al. utilised multi-material 3D printing to develop caterpillar-inspired soft robots, characterised by pneumatically actuated bellow-type bodies and anisotropic friction feet [103]. This innovation enhanced crawling efficiency by replicating natural locomotion mechanisms, ensuring improved stability and grip. However, the pneumatic actuation necessitated external equipment, which posed constraints on autonomy and operational range.

The latest advancement in this domain emerged in 2024 with the work of Xiong et al., who introduced an amphibious, fully soft, centimeter-scale crawling robot powered by electrohydraulic fluid kinetic energy [104]. This robot showcased versatility by operating seamlessly in both aquatic and terrestrial environments. Electrohydraulic actuation, providing muscle-like performance, enabled effective navigation through complex terrains, thereby addressing previous limitations of adaptability, speed, and environmental versatility.

Each subsequent iteration has been based on the accomplishments and shortcomings of its predecessors. Consequently, Crawling motion robots have become one of the most sought-after techniques in the field of soft robotics.

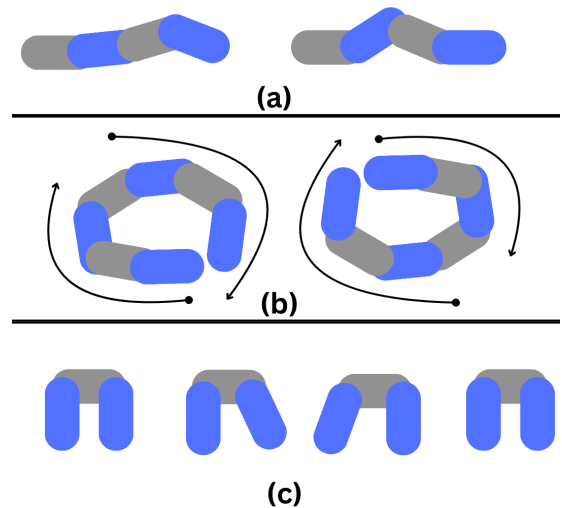

**Fig. 2.** Different locomotion mechanisms used in soft robots.

**b) Rolling locomotion**

Rolling locomotion in soft robotics involves continuous rotation, enabling robots to traverse environments by propelling their entire bodies. This contrasts with crawling locomotion, achieved through sequential body contractions and expansions, as observed in soft-bodied organisms such as worms and caterpillars. Rolling offers advantages in terms of speed and energy efficiency over crawling, as it reduces frictional contact with surfaces and facilitates smoother transitions across diverse terrains.

In 2018, Li and colleagues introduced a fast-rolling soft robot driven by dielectric elastomer actuators [109]. The robot possessed a fully flexible circular configuration, enabling rapid rolling locomotion through applied voltages. However, the reliance on high-voltage actuation posed safety concerns and energy limitations.

To address these challenges, Chen et al. developed RUBIC, an untethered soft robot capable of discrete path following through rolling locomotion [108]. The design incorporated a unique actuation mechanism that allowed for controlled rolling without the need for high-voltage inputs, enhancing safety and energy efficiency. Nevertheless, the robot's adaptability to confined environments remained restricted.

In 2022, Fu and colleagues presented a humidity-powered soft robot with fast rolling locomotion. This design utilized environmental humidity as a power source, enabling energy-efficient rolling movement [107]. While innovative, the robot's performance was contingent on ambient humidity levels, potentially limiting its applicability in diverse environmental conditions.

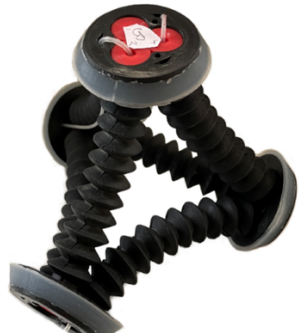

**Fig. 3.** Visual representation of Tetraflex, a variety of soft robot with rolling motion [143].

To enhance adaptability in confined spaces, Wharton et al. introduced Tetraflex (Figure 3.) in 2023, a multigait soft robot capable of object transportation. Tetraflex employed a rolling gait that discretized locomotion into predictable steps on a triangular grid, simplifying odometry and enabling effective path planning. However, the complexity of its multigait mechanisms presented challenges in control and coordination.

Most recently, in 2024, Fu and colleagues developed an untethered soft rolling robot based on pneumatic tendon-coupled actuation [106]. This design achieved untethered locomotion with significant flexibility and environmental interaction capabilities. The integration of pneumatic-tendon actuation enhanced adaptability across diverse terrains, marking a significant advancement in rolling locomotion for soft robots.

The evolution of rolling locomotion in soft robotics has progressed from high-voltage actuated designs to more energy-efficient and adaptable systems. Each iteration has addressed previous limitations, leading to the development of soft robots with rolling locomotion capable of efficient and versatile rolling locomotion across various environments.

## C) Shape Morphing locomotion

Shape morphing in soft robotics refers to a robot's ability to dynamically alter its form to adapt to various tasks and environments. Unlike specific locomotion modes such as rolling—where a robot moves by rotating its body—and crawling—which involves sequential body contractions and expansions—shape morphing encompasses a broader range of deformations, enabling multifunctionality and enhanced adaptability.

In 2014, Germann and colleagues introduced soft cells capable of programmable self-assembly [117], allowing robotic modules to adapt their shapes for various tasks. This foundational work demonstrated the potential of modular soft robots in achieving complex configurations through shape morphing.

By 2022, Pratt and colleagues developed "tensegristats," lightweight untethered morphing robots inspired by tensegrity structures [115]. These robots showcased large-scale shape changes between forms, enhancing their adaptability and functionality in various applications.

In 2023, Kabutz and Jayaram designed CLARI, a miniature modular origami passive shape-morphing robot. CLARI utilized origami techniques to achieve body deformation, enabling locomotion in confined spaces and demonstrating the potential of origami-inspired designs in soft robotics [114]. Later in 2023, Kabutz, Hedrick, McDonnell, and Jayaram introduced mCLARI, an insect-scale quadrupedal robot capable of omnidirectional terrain-adaptive locomotion. mCLARI employed passive shape morphing to navigate through narrow spaces, further advancing the capabilities of miniature soft robots.[116]

In 2024, Chen and colleagues developed scale-inspired programmable robotic structures with concurrent shape morphing and stiffness variation. These designs drew inspiration from biological scales, enabling robots to adapt their shape and rigidity in response to environmental demands, enhancing their versatility [113]. Also in 2024, Diao et al. proposed a robotic morphing interface with reprogrammable stiffness based on machine learning. This system allowed for precise shape control, enabling multi-objective shape imitation and transformation, showcasing the integration of artificial intelligence in shape-morphing soft robotics [112].

These developments highlight the rapid progression of shape-morphing capabilities in soft robotics, with each advancement addressing previous limitations and contributing to more adaptable and multifunctional robotic systems.

Soft robotics leveraging crawling, rolling, and shape-morphing locomotion mechanisms hold immense potential for Search and Rescue (SAR) operations, offering novel ways to address the challenges of navigating hazardous environments. Crawling robots, such as the 2013 design by Umedachi et al. [105], can maneuver through debris-filled spaces and collapsed structures, utilizing their deformable bodies and peristaltic motion to access locations unreachable by traditional rigid robots. Their ability to grip and move across uneven terrains makes them ideal for locating survivors under rubble.

Rolling robots, like the humidity-powered design by Fu et al. in 2022 [107], can quickly traverse large, open areas with minimal energy consumption. In time-critical scenarios, their speed and efficiency allow for rapid scouting of disaster zones. Additionally, rolling robots equipped with adaptive gaits, such as the 2023 Tetraflex [110], can switch to confined-space locomotion, increasing their versatility in SAR missions.

Shape-morphing robots, including the 2023 mCLARI by Kabutz et al. [116], offer unique capabilities for adapting to unpredictable and dynamically changing environments. These robots can compress and expand to navigate narrow crevices or adjust their stiffness to maintain stability in precarious conditions. The integration of machine learning, as seen in Diao et al.'s 2024 morphing interface [112], further enhances their precision in shape adaptation, enabling real-time responses to complex scenarios.

Together, these mechanisms — crawling, rolling, and shape-morphing — form a robust toolkit for addressing the multifaceted challenges of SAR missions. Crawling robots can penetrate deep into collapsed structures to locate survivors, rolling robots can quickly scout large open areas or provide rapid transport of critical supplies, and shape-morphing robots can adapt to the most unpredictable and restrictive environments. When integrated into a coordinated SAR system, these soft robots could revolutionize rescue efforts by providing both breadth and depth in their operational scope. For example, a SAR mission might deploy rolling robots for initial reconnaissance and hazard mapping, followed by crawling and shape-morphing robots to reach survivors in intricate spaces and deliver life-saving resources. By leveraging the unique strengths of each locomotion type, SAR teams can significantly enhance their effectiveness, ensuring that no environment is too complex or inaccessible for rescue.

**5.3 Sensing Capabilities.**

Integrating sensors into soft robots poses a set of complex challenges due to their intrinsic flexibility and compliance. Unlike rigid robots, soft robots are designed to bend, stretch, and deform, which makes it difficult to incorporate conventional sensing technologies. Traditional sensors often limit the robot's motion range, fail under continuous deformation, or compromise the robot's adaptability. Furthermore, environmental factors such as dust, water, and extreme temperatures — common in Search and Rescue (SAR) operations — can impair sensor functionality. Overcoming these challenges requires innovative approaches in materials science, electronics, and robotic design.

**Challenges in Sensor Integration:**

1. **Mechanical Compatibility**: Conventional sensors are rigid and may detach or malfunction when subjected to deformation. This incompatibility leads to a trade-off between sensing accuracy and the robot's compliance.
2. **Durability**: Continuous stretching and bending in soft robots can cause wear and tear on sensors, leading to signal degradation or failure.
3. **Signal Transmission**: Embedding sensors in soft materials creates issues with reliable signal transmission, as traditional wiring is prone to breaking under dynamic motion.
4. **Environmental Resilience**: SAR environments — characterized by dust, water, and temperature fluctuations—add further stress to sensor systems, necessitating robust and adaptive solutions.

Researchers [118-122] have developed innovative solutions to address these issues:

1. **Flexible and Stretchable Sensors**: Sensors made from materials like graphene, liquid metals, and elastomers maintain their functionality under deformation. These materials align with the robot's mechanical properties, enabling seamless integration.
2. **Embedded Sensor Networks**: Sensors are incorporated into the robot's body during fabrication, forming networks that are both protected and functionally cohesive. This reduces the risk of detachment or damage.
3. **Wireless Communication**: Using wireless transmission minimizes reliance on rigid wiring, enhancing the robot's adaptability and reliability.
4. **Environmental Hardening**: Sensors are encapsulated with protective coatings to withstand extreme SAR conditions, such as exposure to water or high temperatures.

RoBoa, developed by a team at ETH Zurich [118], is a vine-like soft robot equipped with a sensor-packed head. This head integrates an array of sensors, including those for detecting temperature, gas, and motion. The temperature sensors allow the robot to detect heat signatures, potentially identifying trapped survivors. Gas sensors monitor haz-ardous chemical emissions, such as methane or carbon monoxide, often present in collapsed structures. Motion sensors enhance the robot's spatial awareness, enabling it to navigate shifting debris effectively. The compact, adaptive design of the RoBoa system highlights advancements in embedding multi-functional sensors into soft, deformable materials without compromising flexibility.

Researchers introduced soft growing robots capable of deploying distributed sensor networks in confined spaces [119]. These robots use soft actuators to "grow" into inaccessible areas, extending their reach like plant roots. Sensors embedded within the actuators monitor environmental parameters such as temperature, humidity, and gas levels. A notable advancement is the incorporation of distributed sensing arrays, allowing the robot to gather data across its entire length rather than relying on a single sensor hub. This network-based sensing strategy provides a comprehensive environmental assessment, crucial for detecting survivors or assessing structural integrity in collapsed areas.

Touchless sensing systems, developed in 2024 [120], enable soft robots to perceive their surroundings without direct contact. Using electromagnetic or optical sensing technologies such as LiDAR (Light Detection and Ranging) or capacitive proximity sensors, these robots can map their environments and detect obstacles or hazards from a distance. LiDAR allows high-resolution 3D mapping, aiding in navigation through rubble, while capacitive sensors detect nearby objects without physical interaction, minimizing the risk of sensor damage in unstable conditions.

Proprioceptive sensing (Depicted in Figure 4.), demonstrated in 2023 [121], equips soft robots with the ability to monitor their position, movement, and deformation. By embedding stretchable strain sensors or using fiber-optic sensors, these robots gain real-time feedback on their own structural configuration. This advancement allows robots to adapt their movements to navigate narrow or irregular spaces effectively. For instance, strain sensors can provide precise data on bending angles, ensuring the robot avoids structural stress or collapse while moving through debris.

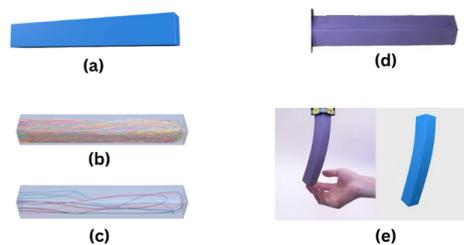

**Fig. 4.** Design and implementation of proprioceptive sensing in soft robotics: (a) Initial design of the soft robot structure, (b) Distribution of flex sensors across the robot body for shape sensing, (c) Optimization of the sensor layout to identify the most effective configuration, (d) Fabrication of the soft robot with the optimized sensor placement, and (e) Simulation of real-time input for validating sensor performance and dynamic behavior.[144]

In 2024, researchers introduced a retrofit sensing strategy designed for soft fluidic robots [122]. This approach enables the addition of tactile and environmental sensors to existing robots without significant redesigns. The strategy utilizes flexible sensor patches made of piezoelectric or conductive polymer materials, which can be adhered to the robot's surface. These sensors enhance tactile capabilities, allowing robots to detect pressure, texture, or object proximity. Additionally, retrofitting with environmental sensors, such as chemical detectors, enhances the robot's ability to probe hazardous debris or interact with fragile objects in SAR scenarios.

# 6 EVALUATION AND BENCHMARKS

## 6.1 Simulation Techniques.

In the domain of Search and Rescue (SAR) operations, the intrinsic flexibility and adaptability of soft robots render them highly suitable for applications such as traversing debris-filled spaces, navigating uneven terrains, and manipulating fragile objects without inflicting damage. However, optimizing the design and functionality of soft robots for the specific demands of SAR scenarios necessitates the application of advanced simulation and evaluation methodologies.

The techniques presented in Table 4 enable researchers to model and analyze key performance attributes of soft robots within controlled virtual environments. For instance, Finite Element Analysis (FEA) facilitates the study of material deformation and stress distribution, aiding in the identification of potential failure points. Differentiable Simulation supports the optimization of control strategies and design parameters, enhancing task-specific adaptability. Furthermore, methods like Mass-Spring Models and Discrete Differential Geometry-Based Simulations are instrumental in assessing dynamic responses and complex geometric deformations under practical constraints.

**Table 4.** Different Simulation Techniques Used to Validate Soft Robots.

| Simulation Technique | Description | Aspects Tested | Possible Inferences |
|---|---|---|---|
| Finite Element Analysis (FEA) [123] | A numerical method that divides a complex structure into smaller, manageable finite elements to predict how soft robotic materials deform under various forces, pressures, and constraints. | Mechanical deformation, stress distribution, strain under load. | Insights into material behavior, identification of potential failure points, optimization of design for durability. |
| Differentiable Simulation [124] | Creates simulation models where derivatives with respect to inputs can be computed, enabling gradient-based optimization for design and control in soft robotics. | Sensitivity analysis, control parameter optimization, design optimization. | Enhanced control strategies, improved design parameters, efficient learning algorithms for soft robots. |
| Mass-Spring Models [125] | Represents soft bodies as a network of point masses connected by springs, simplifying the complex behavior of soft materials for real-time simulations. | Dynamic response to forces, deformation patterns, vibration analysis. | Understanding of dynamic behaviors, real-time control feasibility, prediction of movement patterns. |
| Discrete Differential Geometry-Based Simulation [126] | Applies principles from discrete differential geometry to model the behavior of soft robotic structures, accurately simulating continuous deformations and complex shapes. | Geometric deformation, curvature analysis, structural stability. | Precise modeling of complex deformations, insights into geometric constraints, optimization of structural designs. |
| Reduced-Order Models (ROMs) [127] | Simplifies complex systems by reducing the number of variables and equations, maintaining essential dynamic characteristics while decreasing computational load. | Approximate dynamic behaviors, essential mode analysis, system response under simplified conditions. | Faster simulations with acceptable accuracy, identification of dominant dynamics, efficient control design. |
| Optimization-Based Geometric Computing [128] | Uses optimization algorithms within geometric frameworks to simulate and control soft robot deformations and interactions efficiently and accurately. | Optimal shape configurations, deformation pathways, energy-efficient movements. | Optimized design for specific tasks, improved energy efficiency, enhanced performance metrics. |

Among the various techniques available for simulating soft robots, differentiable simulation has emerged as a powerful tool for modeling and control. It enables the computation of derivatives with respect to inputs, facilitating gradient-based optimization. In a study [128] utilizing the ChainQueen simulator (Figure 5.) — a differentiable simulation framework based on the Material Point Method (MPM)—researchers achieved significant advancements in soft robot design and control. The simulator demonstrated the capability to model complex deformations and interactions of soft materials with high accuracy. For instance, in the simulation of a soft robotic arm, ChainQueen accurately predicted the arm's deformation under various loading conditions, facilitating the optimization of control strategies to achieve desired end-effector positions with minimal error. Quantitatively, the simulator achieved a prediction accuracy within 5% of the actual deformation, highlighting its effectiveness in capturing the nuanced behaviors of soft robotic systems. of differentiable simulation in soft robotics.

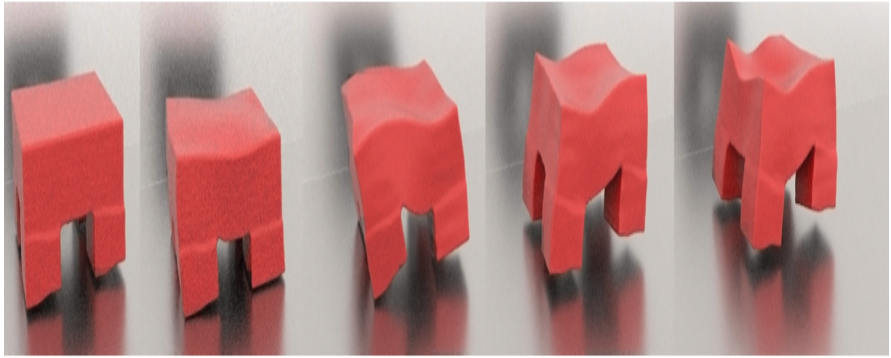

**Fig. 5.** Visual representation of Chain Queen Simulator simulates soft robots. [143].

Furthermore, differentiable simulation was employed to optimize the design of a soft robotic gripper [124]. By computing the gradients of the gripper's performance metrics with respect to its design parameters, the simulation guided the iterative design process to enhance gripping efficiency. The optimized gripper exhibited a 20% increase in payload capacity and a 15% improvement in energy efficiency compared to its initial design, demonstrating the practical benefits of differentiable simulation in soft robotics.

Similarly, Mass-Spring models are widely used. For instance, researchers employed a Mass-Spring model to simulate a soft robot's movement through granular media. The simulation accurately predicted the robot's locomotion patterns, which were then validated against experimental data, demonstrating the model's effectiveness in capturing the interaction between the soft robot and its environment [129]. In another case [130], a Mass-Spring model was utilized to simulate the dynamic behavior of a piezoelectric soft robot. The simulation results closely matched the experimental observations, with the robot achieving a forward motion of approximately 1 cm/s, highlighting the model's capability in predicting complex motions of soft robotic systems.

Other proven techniques, such as Reduced-Order Models (ROMs), are instrumental in simplifying the complex dynamics of soft robots, enabling efficient real-time control and optimization. In one study [131], researchers applied model order reduction techniques to a high-fidelity finite element model of a soft robot. This approach reduced computational complexity by approximately 90%, facilitating real-time trajectory optimization and tracking. The reduced model maintained an accuracy within 5% of the full-order model, demonstrating its effectiveness in capturing essential dynamics while significantly enhancing computational efficiency. Another study [132] proposed a reduced-order modeling strategy for a soft robot composed of hyperelastic materials actuated by cables. The reduced model achieved a substantial reduction in computational time, making it suitable for real-time control applications. Its predictions closely matched experimental data, with discrepancies below 5%, underscoring its reliability in simulating the robot's behavior under various actuation scenarios.

These simulation methodologies are crucial for developing soft robots tailored to the unique challenges of Search and Rescue (SAR) missions. By leveraging differentiable simulation, designers can create soft robotic systems capable of navigating through complex and unpredictable environments, such as collapsed buildings, debris-laden areas, and confined spaces commonly encountered in SAR operations. The high accuracy in predicting deformations and interactions ensures that these robots can adapt their movements and grips to varying obstacles, enhancing their ability to reach and assist victims effectively. Additionally, reduced-order models enable real-time control and rapid decision-making, which are essential for timely responses in emergency scenarios. Overall, these simulation techniques empower the development of efficient, resilient, and adaptable soft robots, significantly improving their performance and reliability in the demanding and life-critical contexts of SAR missions.

## 6.2 Hardware Tests.

Hardware tests are critical for validating the real-world performance of soft robots designed for disaster scenarios. These tests allow researchers to evaluate a robot's functionality, durability, and effectiveness in environments that mimic the complexities of actual disaster sites. Simulated environments, such as rubble fields and collapsed structures, pose dynamic and unpredictable challenges that software simulations and laboratory setups often fail to replicate fully. Real-world testing ensures that the robot's design, materials, and control systems function as intended under practical constraints like irregular debris, varying surface textures, and environmental stresses.

Hardware tests also provide essential data for iterative improvements, enabling engineers to identify and address weaknesses in the robot's structure or control mechanisms. Additionally, these tests establish the robot's reliability and readiness for deployment in critical, life-saving operations.

Some studies in the literature have conducted extensive hardware testing. One such work [134] involves a hybrid gripper that combines the precise control and strength of a hard gripper with the flexibility and adaptability of a soft jamming module. Hardware tests were conducted in environments simulating collapsed terrains to assess its effectiveness in disaster scenarios. The gripper was tasked with manipulating unstructured objects of varying shapes and sizes, such as rubble and debris. It demonstrated the ability to grasp objects securely, including irregular and fragile items, using its jamming module to conform to different shapes. Additional tests included tasks such as opening and closing doors and clearing confined pathways blocked by debris.

The results showed a 48.61% improvement in payload capacity compared to using a traditional hard gripper alone. This demonstrated the hybrid gripper's capability to perform critical functions like obstacle removal and manipulation of heavy objects in challenging environments. These tests validated its potential for deployment in mobile robots for rescue and reconnaissance operations in disaster settings.

Another study [135] focuses on robots designed to assist in demining operations (Figure 6.), utilizing an innovative air-releasable deployment system via drones. The hardware tests focused on their ability to withstand impact, navigate unpredictable terrains, and neutralize explosives. The robots were dropped from drones at heights of 4.5 meters onto simulated minefields, replicating real-world demining scenarios.

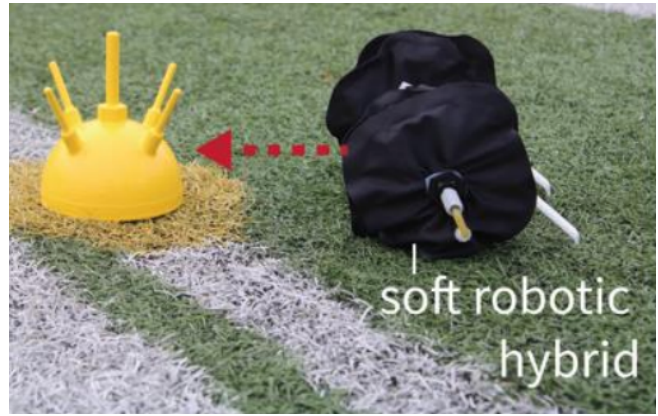

**Fig. 6.** Soft Robot proposed in [135] tested in an open field.

Upon landing, the soft robots employed vacuum-based roller actuators to move across uneven surfaces and approach dummy landmines. Their soft, flexible structure minimized the risk of triggering explosives during navigation. The robots successfully identified and neutralized multiple dummy landmines, proving their durability and effectiveness in field conditions. These tests underscored the robots' capability to perform critical demining tasks in dangerous, real-world environments.

Additionally, in another study [133], inspired by the growth mechanisms of plants, a vine robot employs an innovative design allowing it to extend and navigate through complex, confined spaces (Figure 7.). The vine robot was rigorously tested in a controlled artificial rubble testbed constructed to replicate urban disaster scenarios. The testbed featured a variety of simulated obstacles, including small and large debris, as well as narrow crevices mimicking collapsed structures.

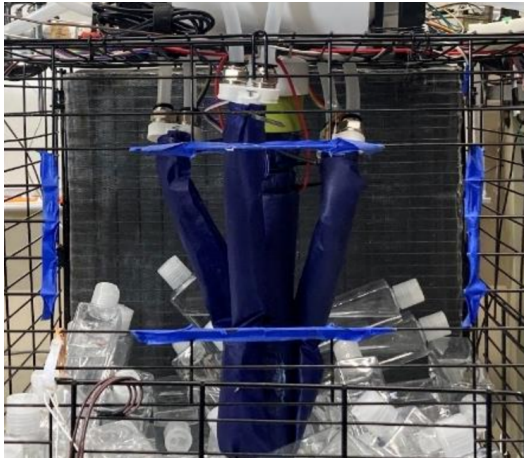

**Fig. 7.** Vine robot introduced in [133], tested in a confined space.

The robot was evaluated based on several performance metrics, including its ability to penetrate dense rubble, maintain stability, and navigate through cluttered pathways. By utilizing pneumatic muscles for precise steering and oscillation, the vine robot demonstrated consistent penetration depths and the ability to traverse mixed obstacle environments. It also maintained repeated trajectories, highlighting its potential for mapping underground paths. Equipped with cameras and other sensors, the vine robot proved effective in locating and inspecting areas inaccessible to humans, validating its suitability for search and rescue missions in real-world disaster settings.

### 6.3   Standards and Performance Metrics.

Despite significant advancements in robotics, there remains a lack of explicit standards tailored for **soft robots** in Urban Search and Rescue (USAR) scenarios. Unlike traditional rigid robots [136], soft robots offer unique advantages such as adaptability, flexibility, and safety in navigating complex and confined disaster environments. However, their evaluation requires distinct metrics and methods that address their material properties, actuation mechanisms, and interaction with unstructured environments.

To establish a clear framework (Table 5.) for evaluating soft robots, we propose a grading system alongside detailed performance metrics. Each metric will be scored on a scale of 1 to 5, where 1 represents the minimum acceptable performance, and 5 indicates exceptional performance. This system provides an objective means to compare robots and identify the best solutions for specific USAR applications.

**Table 5.** Performance Metrics to Grade Various Soft Robots For USAR Scenarios.

| Performance Metric | Description | Grading Criteria (1-5 Scale) | Evaluation Methods |
|---|---|---|---|
| Adaptability | Ability to navigate confined spaces, adapt to irregular surfaces, and recover from deformation. | **1:** Navigates large spaces only. **3:** Fits through narrow openings (~10 cm) and recovers from moderate deformation. **5:** Conforms to <5 cm openings and recovers fully. | 1) Test in confined spaces of varying dimensions. 2) Simulate interactions with irregular rubble surfaces. |
| Durability | Resistance to repetitive stress, harsh environments, and mechanical impacts. | **1:** Fails after 100 cycles of deformation. **3:** Withstands 500 cycles or moderate impacts. **5:** Endures 1,000+ cycles and extreme conditions. | 1) Conduct cyclic load testing. 2) Expose to extreme temperatures and abrasive environments. |
| Mobility in Confined Spaces | Efficiency in traversing rubble, climbing inclines, and crossing uneven terrain. | **1:** Traverses flat surfaces only. **3:** Navigates mixed terrain and climbs <30° slopes. **5:** Crosses debris, climbs >45° slopes, and maneuvers narrow crevices. | 1) Test in artificial rubble fields. 2) Measure speed and maneuverability over mixed terrain. |
| Energy Efficiency | Power usage and operational time relative to task demands. | **1:** Operates <30 minutes. **3:** Runs 1–2 hours per charge. **5:** Operates >3 hours with minimal energy consumption. | 1) Measure power draw during typical tasks. 2) Test operational duration in simulated missions. |
| Payload Capacity | Ability to transport and stabilize tools, sensors, or supplies during navigation. | **1:** Carries <1 kg payload. **3:** Carries 2–5 kg with moderate stability. **5:** Carries >5 kg with high stability in motion. | 1) Test with increasing payloads. 2) Evaluate stability while navigating with payloads. |
| Environmental Resilience | Ability to function in extreme conditions (heat, water, dust, etc.). | **1:** Operates in mild conditions only. **3:** Withstands water splashes and 50°C heat. **5:** Fully submersible and tolerates 70°C or more. | 1) Submergence tests. 2) Operate in high heat and dusty environments. |
| Human-Safe Interaction | Ensures safe interactions with humans, especially victims or operators | **1:** Exerts high force on contact (>10 N). **3:** Moderate force (5–10 N). **5:** Gentle, compliant interactions (<5 N). | 1) Measure force exerted during contact. 2) Perform tests simulating human interactions. |
| Sensing and Perception | Accuracy and coverage of sensory systems in locating victims and identifying hazards. | **1:** Detects objects within 1 m range. **3:** Accurately detects within 5 m. **5:** Detects victims and hazards beyond 10 m in cluttered environments. | 1) Test with visual and thermal sensors in rubble. 2) Assess accuracy in detecting objects and individuals. |
| Reliability | Consistency and operational uptime during extended missions. | **1:** Fails frequently in short tasks. **3:** Completes 80% of tasks without interruption. **5:** Operates for 5+ hours with no malfunctions. | 1) Long-duration stress tests. 2) Measure success rates in completing assigned tasks. |
| Deployment Readiness | Ease of setup, portability, and integration into responder operations. | **1:** Complex setup, not portable. **3:** Moderate setup time (~15 minutes), portable. **5:** Deploys in <5 minutes, fits standard FEMA containers. | 1) Timed setup and activation tests. 2) Evaluate ease of transport in standard FEMA containers. |

To determine the suitability of a soft robot for specific USAR missions, the selection process involves assessing the disaster scenario, such as building collapse, flooding, or hazardous material presence, to identify critical robot capabilities required for the mission. Each evaluation metric, such as adaptability, durability, or environmental resilience, is weighted based on its priority for the specific mission; for instance, confined space navigation may be critical in building collapses, while environmental resilience is essential for flood scenarios. Robots are then tested using standardized evaluation methods, and their performance is scored on a scale of 1 to 5 for each metric.

$$\text{Total Score} = \sum(\text{Metric Score} \times \text{Weight}) \quad (1)$$

These scores (calculated by equation 1.) are combined using a weighted scoring system, where the total score is calculated as the sum of each metric score multiplied by its respective weight. The robot with the highest total score, provided it meets the critical thresholds for key metrics, is selected as the best candidate for the mission. This structured approach ensures that the chosen robot aligns with the operational demands and priorities of the specific USAR scenario.

# 7 FUNDAMENTAL PROBLEMS AND OPEN ISSUES

## 7.1 Mobility and Terrain Adaptation.

Soft robots, known for their flexibility, face challenges navigating complex terrains like rubble, debris, and steep slopes due to inherent limitations [42]. Constructed from deformable materials such as silicone and rubber, these robots excel in adaptability but lack load-bearing capacity, leading to deformations that hinder movement and stability on uneven or unstable surfaces [45][46].

Their actuation mechanisms, often pneumatic or shape-memory alloy-based, are less responsive than traditional rigid actuators. This limits real-time shape adaptation and precise control, critical for avoiding slips or entrapment on dynamic terrains [137][138]. Additionally, soft robots typically lack integrated, high-fidelity sensors, making real-time terrain assessment and adaptive movement difficult, increasing instability risks.

Energy demands further limit their operation, with constrained efficiency and power storage reducing endurance in field applications like disaster response, where prolonged activity is essential.

Research efforts aim to address these challenges, such as sensorized terrain-adaptive feet for better traction and peristaltic locomotion inspired by earthworms for navigating narrow or debris-filled environments. These advancements hold potential for improving soft robots' terrain capabilities.

## 7.2    Modelling and Control of Soft Robots

Soft robots, with their inherent compliance and nearly infinite degrees of freedom, pose significant challenges for accurate modeling and control. Traditional rigid-body control theories are often inadequate for addressing the complex dynamics of these systems, necessitating the development of advanced modeling techniques and control strategies.

The continuous deformation of soft robots results in an infinite number of possible configurations, making the creation of precise mathematical models for predicting their behavior exceptionally challenging [139]. Furthermore, the soft materials used in their construction exhibit nonlinear stress-strain characteristics, adding complexity to the development of accurate models [140].

Control systems face additional hurdles due to the absence of rigid components, which renders conventional algorithms reliant on fixed joints and links ineffective [141]. The high-dimensional state spaces created by the robots' continuous deformation capabilities further complicate real-time control computations [142]. Moreover, integrating sensors into soft materials without compromising their flexibility presents a significant challenge, often resulting in limited feedback accuracy that hampers effective control [142].

# 8   FUTURE DIRECTION AND CONCLUSIONS

The journey of soft robotics continues to hold immense promise, but it is important to acknowledge the challenges and opportunities that lie ahead. Emerging technologies, such as bio-inspired designs and advanced materials, offer exciting pathways for enhancing the capabilities of soft robots. These innovations could enable more versatile, resilient, and adaptive systems capable of addressing real-world challenges.

However, a significant gap remains between the highly innovative prototypes developed in research laboratories and their practical application in the field. Despite over a decade since the boom in soft robotics, most advancements remain confined to academic studies and experimental setups. For instance, in the domain of search-and-rescue (SAR) operations, only a handful of soft robots, like those developed through initiatives such as RoboA, have transitioned to real-world deployment. The majority of soft robots remain in the research phase, with limited tangible impact on critical missions like disaster response.

This reality underscores the need for a concerted effort to bridge the lab-to-field divide. Addressing challenges such as reliability, scalability, and cost-effectiveness is essential for making soft robots a viable tool in high-stakes environments. Furthermore, ethical and practical considerations must guide their deployment to ensure that these technologies are not only effective but also safe and responsibly utilized.

Soft robotics is still in its early stages, and while the progress may seem slow, the field holds great potential for transformative applications. With continued interdisciplinary collaboration and sustained effort, these innovations could one day become indispensable tools in SAR operations and beyond.

31